\documentclass[sn-basic]{sn-jnl}

\jyear{2021}%

\theoremstyle{thmstyleone}%
\theoremstyle{thmstyletwo}%

\theoremstyle{thmstylethree}%

\usepackage{tabularx}
\usepackage{graphicx}
\usepackage{adjustbox}
\usepackage{bbding}
\usepackage{subfig}
\renewcommand{\thetable}{\arabic{table}}
\begin{document}

\title{Vertical Federated Learning: A Structured Literature Review}


\author*[1]{\fnm{Afsana} \sur{Khan}}\email{a.khan@maastrichtuniversity.nl}

\author[1]{\fnm{Marijn} \sur{ten Thij}}\email{m.tenthij@maastrichtuniversity.nl}

\author[1]{\fnm{Anna} \sur{Wilbik}}\email{a.wilbik@maastrichtuniversity.nl}

\affil*[1]{\orgdiv{Department of Advanced Computing Sciences}, \orgname{Maastricht University}, \orgaddress{\street{Paul-Henri Spaaklaan}, \city{Maastricht}, \postcode{6229}, \country{The Netherlands}}}


\abstract{Federated Learning (FL) has emerged as a promising distributed learning paradigm with an added advantage of data privacy. With the growing interest in having collaboration among data owners, FL has gained significant attention of organizations. The idea of FL is to enable collaborating participants train machine learning (ML) models on decentralized data without breaching privacy. In simpler words, federated learning is the approach of ``bringing the model to the data, instead of bringing the data to the model''. Federated learning, when applied to data which is partitioned vertically across participants, is able to build a complete ML model by combining local models trained only using the data with distinct features at the local sites. This architecture of FL is referred to as vertical federated learning (VFL), which differs from the conventional FL on horizontally partitioned data. As VFL is different from conventional FL, it comes with its own issues and challenges. In this paper, we present a structured literature review discussing the state-of-the-art approaches in VFL. Additionally, the literature review highlights the existing solutions to challenges in VFL and provides potential research directions in this domain.}

\keywords{Federated Learning, Vertically Partitioned Data, Privacy-Preserving Machine Learning, Literature Review}



\maketitle

\section{Introduction}\label{sec1}
The use of machine learning (ML) has enabled organizations to more quickly identify potentially profitable opportunities as well as risks that may be involved. As more and more data becomes accessible over time, there has been a corresponding rise in interest in the application of machine learning across a variety of fields. For instance, in healthcare, the use of machine learning to analyze the health data generated by the growing number of wearable devices like smart watches and fit bits is gaining momentum \citep{bhardwaj2017study}. Moreover, the growing use of ML in financial systems has transformed industries and societies. From traditional hedge fund management firms to FinTech service providers, many financial firms are investing in data science and ML expertise \citep{goodell2021artificial}. ML has also made a significant contribution to the agriculture sector by creating new opportunities to unravel, quantify, and understand data intensive processes in agricultural operational environments \citep{liakos2018machine}. 
\\\\
While organizations can benefit from applying machine learning techniques to their own data, doing the same with data from other comparable organizations could result in significant improvements to the existing organizational processes. In order to build sophisticated machine learning models for improving consumer service and acquisition, substantial emphasis has been placed on integrating data from various organizations, indicating the importance of collaboration. However, the traditional approach of bringing data located at different sites into a central server for training machine learning models is not always feasible as it raises numerous concerns. At present, sharing data among organizations has become critical due to concerns about privacy, maintaining competitive advantages, and/or other constraints. Data security and privacy are issues that are being prioritized not just by individuals or organizations but also by the larger society. The General Data Protection Regulations (GDPR), which the European Union put into place on May 25, 2018 \citep{albrecht2016gdpr} aims to protect users' personal privacy and data security. To address this issue, federated learning (FL), a new distributed learning paradigm, has recently received a lot of attention. FL allows collaboration among organizations to train machine learning models while ensuring that private data of these organizations are not disclosed \citep{ma2020safeguarding}. \cite{kairouz2021advances} formally defined federated learning as \\
\begin{center}
``\emph{A machine learning setting where multiple entities (clients) collaborate in solving a machine learning problem, under the coordination of a central server or service provider. Each client’s raw data is stored locally and not exchanged or transferred; instead focused updates intended for immediate aggregation are used to achieve the learning objective}''
\end{center}
\begin{figure*}[ht!]
\centering
\setkeys{Gin}{width=0.5\textwidth}
\subfloat[Horizontal Federated Learning,
\label{hfl}]{\includegraphics{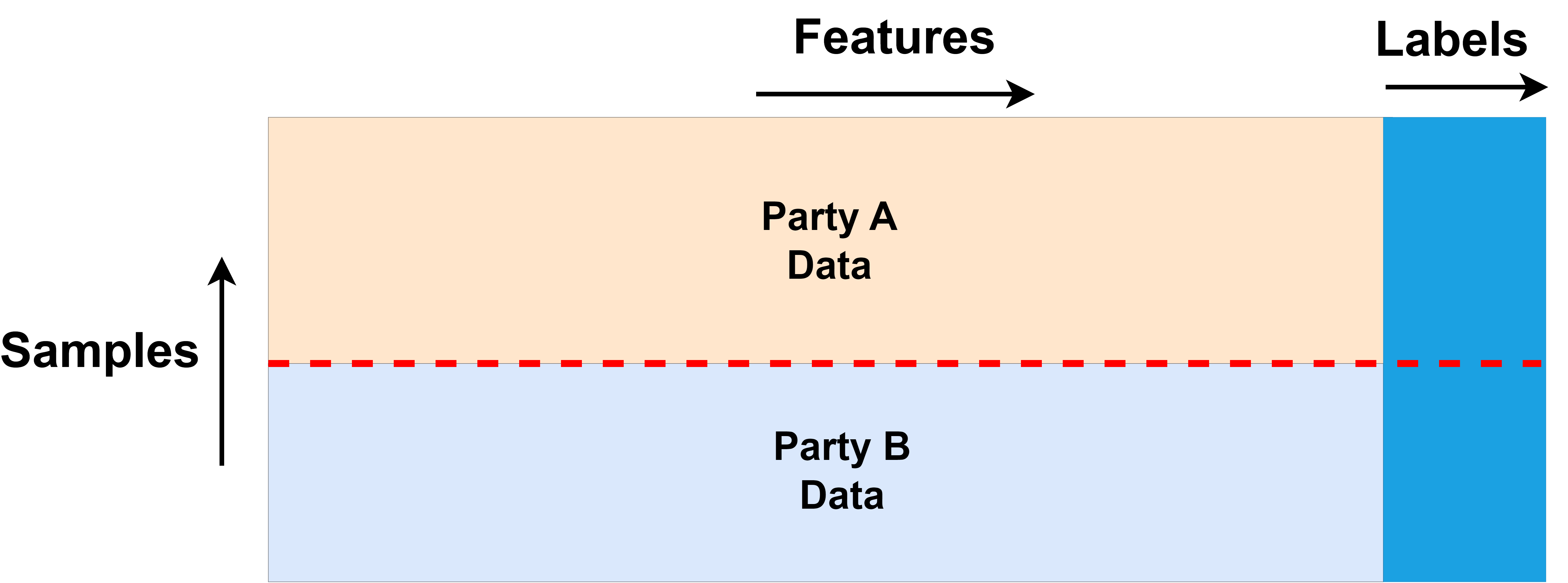}}
\hfill
\subfloat[Vertical Federated Learning,
\label{vfl}]{\includegraphics{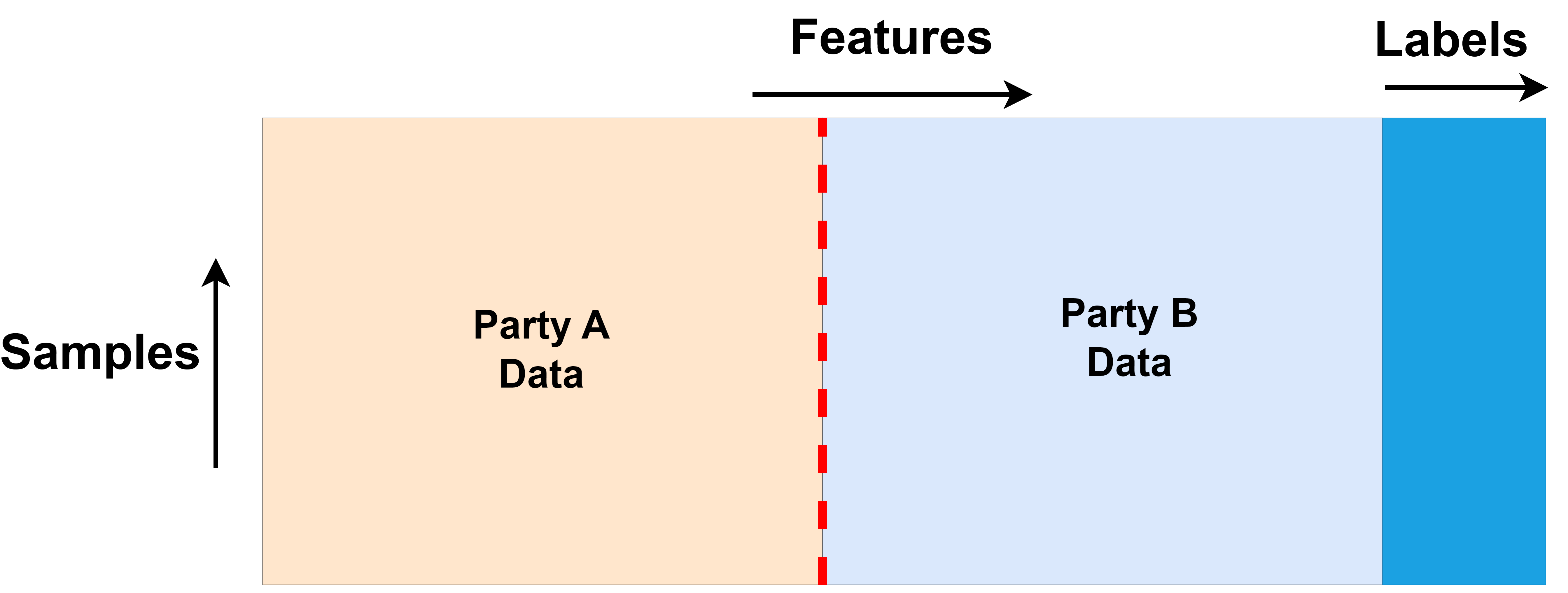}}
\\
\subfloat[Hybrid Federated Learning,
\label{hybrid}]{\includegraphics{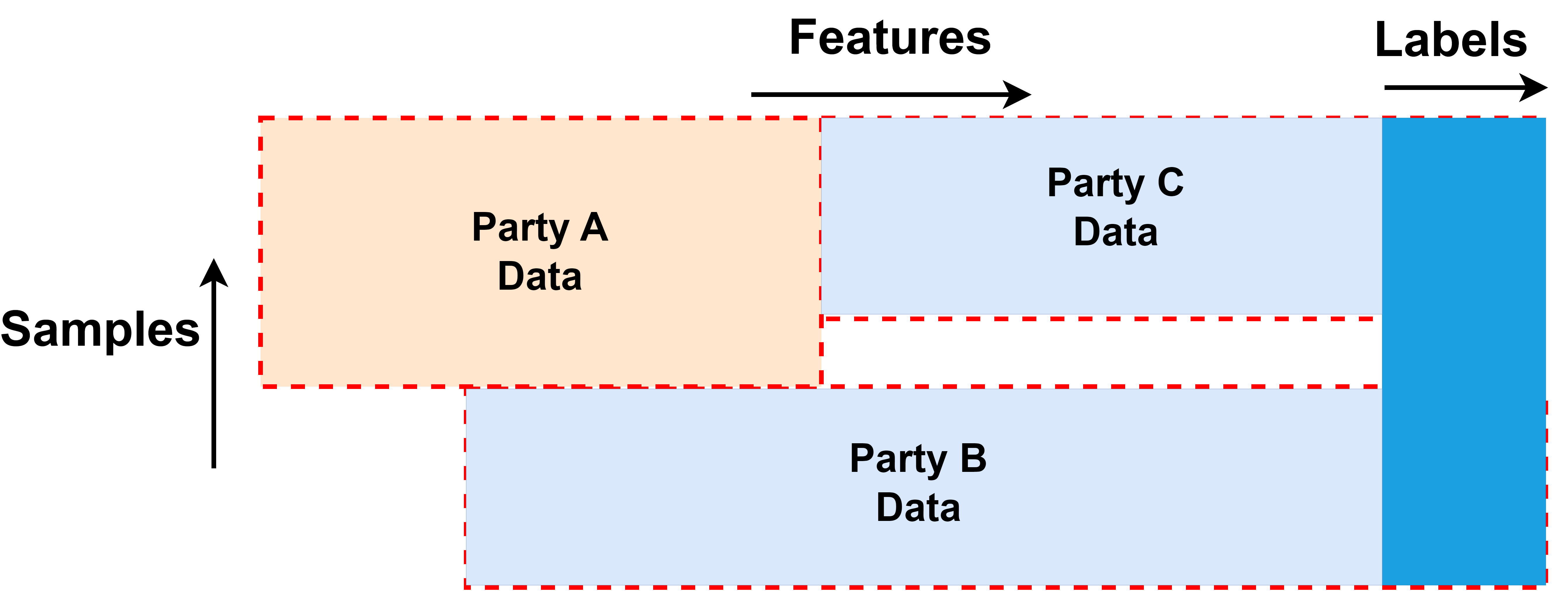}}
\caption{Types of Federated Learning}
\label{paritions}
\end{figure*}
Depending on how the data is partitioned or distributed among organizations, FL can be classified into three scenarios; Horizontal, Vertical and Hybrid. Horizontal federated learning (HFL) is suitable in scenarios when organizations share the same attribute space but differ in samples (Figure \ref{hfl}). An example of HFL is a group of hospitals collaborating to build a ML model used to predict health risks for their patients, based on agreed-upon data. However, HFL sometimes has limited applications in practical scenarios, for example, in fostering collaboration among organizations with competing interests. Due to business reasons, it is more likely that organizations will not be willing to collaborate with their competitors \citep{cheng2020federated}. On the other hand, vertical federated learning (VFL) is suitable for scenarios where organizations have the same set of samples as their data but differ in feature space (Figure \ref{vfl}). VFL promotes collaboration among non-competing organizations with vertically partitioned data. In such cases, typically one organization has the ground truth, or labels, with some of the features of a number of samples. The rest of the participants take part in the federation by providing additional feature information of the same sample space but at the same time ensuring that their data is not disclosed directly to other participants. In return these participants can be compensated with monetary and/or reputational rewards. Examples where VFL is applicable could be a telecom company collaborating with a home entertainment company (cable TV provider) or an airline collaborating with a car rental agency. Hybrid FL refers to the hybrid situation of horizontally and vertically partitioned data (Figure \ref{hybrid}). In this scenario, the data owners hold different attributes for different data instances. However, hybrid FL is not yet explored significantly in the literature.
\\\\
Although VFL is a promising paradigm for privacy-preserving learning, limited research exists which explored the core challenges and methodologies of VFL due to the fact that, FL itself is a comparatively new concept. The pre-print articles \citep{li2023vertical, liu2022vertical} provide a comprehensive overview of VFL, its taxonomy, threats, and prospects, applications in various domains and offers insights into the potential benefits and challenges of using this approach. However, to the best of our knowledge there is no published article yet that provides a structured review on the current research trends on VFL. Following a structured protocol for review ensures that the review is comprehensive, rigorous, and transparent, and that the results are easily accessible and understandable to the intended audience. Considering this as a motivation, in this article, we present a structured literature review (SLR) designed to summarize existing methodologies of VFL as well as to pinpoint potential future directions in order to address the challenges. Our goal for presenting this review is not diving in depth into particular implementations of VFL but rather providing direction for the researchers interested in VFL. The readers would be able to find detailed information on VFL methods and techniques from the articles cited.

\section{Methodology}
The goal of the following structured literature review is to investigate major challenges and existing solutions for vertical federated learning. The study not only provides an overview of the major publications in VFL, but also to identify potential gaps and opportunities for further research. The review was planned, conducted, and reported in accordance with the SLR process proposed by \cite{armitage2008undertaking}. The SLR consists of five main steps including defining research questions, designing search strategy, selecting studies, extracting data and finally synthesis of data. Below we explain these steps in more detail.
\subsection{Research Questions}
Taking into consideration the objective of the review, the following set of research questions were formulated as the initial stage of this SLR.
\begin{itemize}
    \item \textbf{RQ1: }What are the existing methods in VFL and what problems do they tackle? 
    \item \textbf{RQ2: }What are the existing applications of VFL?
    \item \textbf{RQ3: } What are the potential future directions for research in VFL?
\end{itemize}

\begin{table*}[h!]
\begin{adjustbox}{width=\textwidth}
\begin{tabular}{p{3.2cm}p{1cm}p{1.2cm}p{1cm}p{1cm}p{1.2cm}p{1.2cm}}
Search Terms & Google Scholar & Web of Science & IEEE Xplore & arXiv & No. of Articles & Unique Articles \\\hline
“Vertical federated learning” & 141 & 42 & 63 & 57 & 303 & 205 \\
“Vertical" AND “Federated Learning" & 102 & 37 & 88 & 97 & 324 & 219 \\
“Vertical" AND “privacy-preserving federated learning" & 17 & 4 & 7 & 5 & 33 & 23 \\
“Vertical" AND “Heterogeneous Federated Learning" & 6 & 0 & 0 & 0 & 6 & 6 \\\hline
Total & & & & & 453 & 271 \\\hline
\end{tabular}
\end{adjustbox}
\caption{Search Results}
\label{search-table}
\end{table*}
\subsection{Search Strategy}
After the formulation of the research questions, a plan was made to design the search strategy for the SLR. The search strategy includes the initial task of selecting literature databases. Kitchenham et al. listed several high-quality databases for searching research resources \citep{kitchenham2012systematic}. Since VFL is a trending research topic, there are many articles which are pre-prints. We chose to include both published (conference and journal papers) and pre-prints as a part of the SLR. The selected databases were Google Scholar, Web of Science (WoS), IEEE Xplore and arXiv. We experimented with different search terms on the chosen databases. Finally, it was decided that the following search term would yield the most relevant results:
\begin{center}
\emph{(“Vertical federated learning”) OR ("Vertical" AND (“Federated Learning” OR “privacy-preserving federated learning” OR  “Heterogeneous federated learning”))}
\end{center}

\subsection{Study Selection}
The results obtained using the defined search strategy were filtered based on a set of criteria. Only the articles which met the following criteria were considered further for first round screening:
\begin{itemize}
    \item Published after 2015 since the term \emph{"Federated Learning"} was first coined in 2016
    \item Published until 15/03/2023
    \item Written in English language
    \item Availability of full text
    \item Title and abstract specifically mention the focus on vertical federated learning
\end{itemize}
After the initial screening, the chosen articles were checked for redundancy which resulted in 122 unique articles. The unique articles were then investigated by going through the full text. Additionally, we used "snowballing" \citep{streeton2004researching} technique to identify relevant articles. This was done by identifying relevant articles from the reference section of the previously selected articles. Use of the "snowballing" technique lead to finding 14 additional relevant articles which were included for this review. The final articles which are included in this SLR were selected based on the fact that they answered the following questions
\begin{itemize}
    \item Did the article provide an answer to any of the research questions?
    \item Was the article focused on VFL and not general FL?
    \item Was the method proposed in the article evaluated?
    \item Were the experiments and results properly documented?
\end{itemize}
In addition, survey papers were also considered where vertical federated learning had been addressed.
\subsection{Data Extraction}
We extracted data from each of the included papers and organized it in a manner such that we could provide an analysis of the reviewed literature and use them for synthesis in the next section. The data which were extracted from the articles were title \& year of publication, source of publication, research question/problem solved, proposed method, availability of theoretical analysis, dataset evaluated on and model evaluated with.
\subsection{Data Synthesis}
As a last step of the literature review, we performed an analysis of 136 articles relevant to VFL and clustered them based on the problem those have addressed and solved. A detailed review of these articles have been provided in the further section.
 \begin{figure}[h!]
\centering
\includegraphics[width=\textwidth]{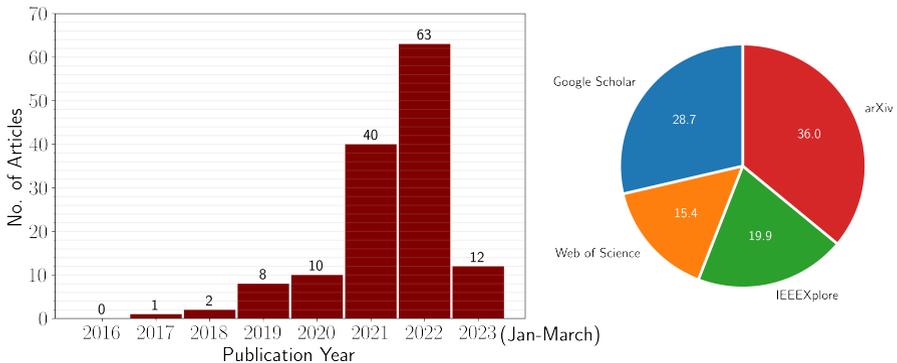}
\caption{Overview of No. of Articles Found; Yearly Trend (Left Panel) \& Databases(Right Panel)}
\label{pub_year}
\end{figure}
 
 \section{Research Results}
\noindent
 We conduct statistical analysis and present the results of a structured literature review by reading and analyzing articles related to FL that are found in the four major databases mentioned earlier. These results of the study provide answers to the research questions that were presented in Section 2.1. To understand the research trend of VFL, we conducted statistics for the publication year of literature as shown in Figure \ref{pub_year}. Although the concept of federated learning was first introduced in 2016, research on FL until 2018 primarily concentrated on horizontal federated learning. But from 2019 a significant increase has been observed in the number of published articles focusing on VFL. Therefore, it can be concluded that VFL is still in its early stages of development.\\


\noindent
Observing from the sources of publications, we found that more than 60\% of the included papers in this literature review are published work in journals and conferences which implies that VFL has produced mature results. Emerging topics frequently have a substantial number of pre-prints available in databases like arXiV. Among the articles studied for this review, 36 percent were pre-prints and available on arXiv. This indicates the rapid development in VFL.
\subsection{Vertical Federated Learning}
\begin{figure}[ht]
\centering
\includegraphics[width=0.6\textwidth]{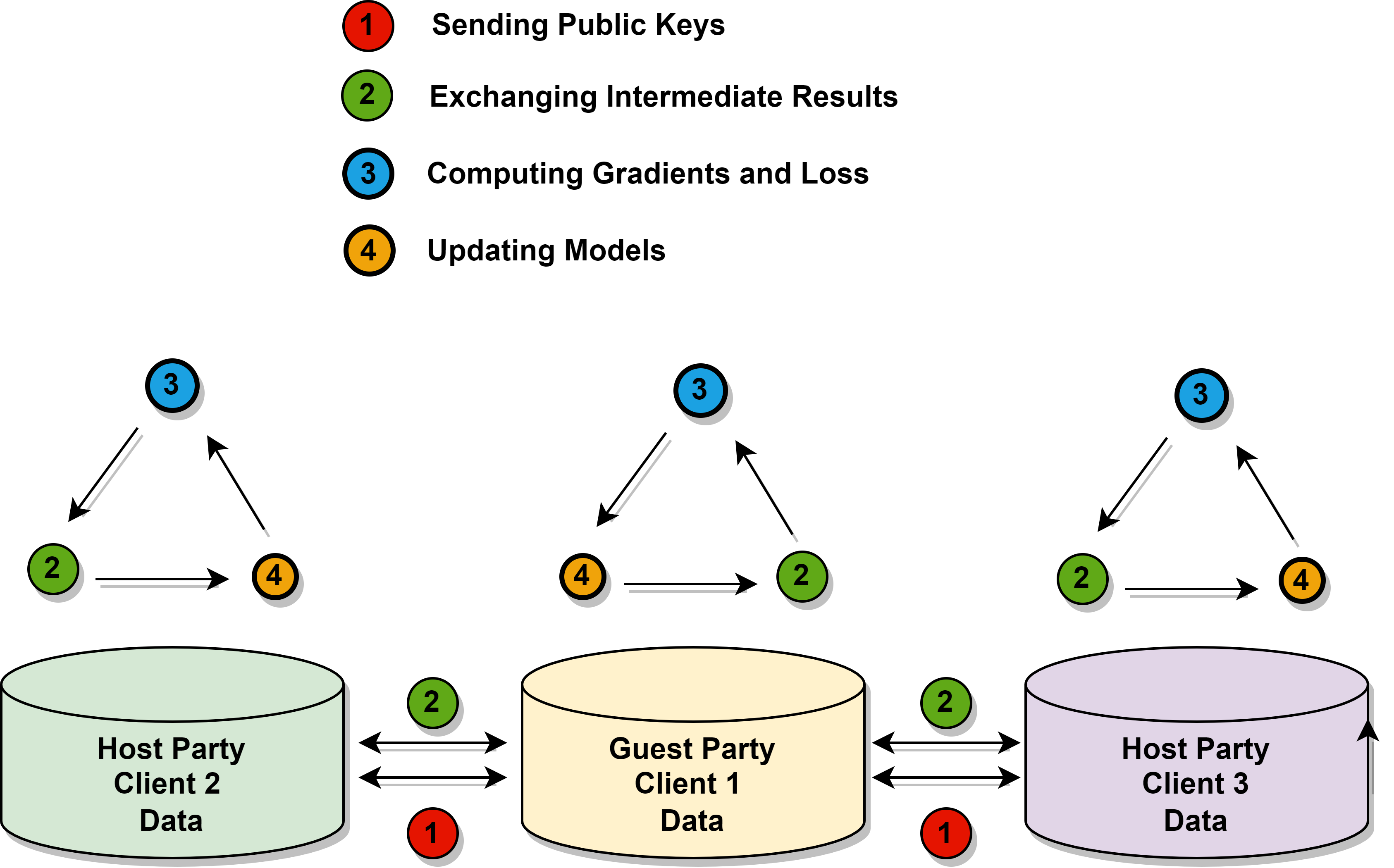}
\caption{Vertical Federated Learning Architecture without Coordinator (Adapted from \citep{yang2019federated})}
\label{architecture}
\end{figure}

In a vertical federated learning setting, the features of data points are distributed among different partitions. There are two basic architectures for VFL; with co-ordinator \citep{hardy2017private} and without co-ordinator \citep{yang2019parallel}. \citep{hardy2017private} proposed a framework comprised of one trusted coordinator and two parties, each of which represents a single client. The task of the coordinator was to  compute training loss as well as generate homomorphic encryption key pairs for privacy. Later on, some research works proposed \citep{yang2019parallel, he2021secure, sun2022privacy} proposed a two-party architecture that eliminated the need for a trusted coordinator, thus reducing the complexity of the system. This architecture was further extended by implementing it in case of multiple collaborating parties/clients \citep{cheng2021secureboost, zhao2022ntp}. Figure \ref{architecture} illustrates the basic protocol for VFL with multiple parties where the party holding the labels is the active party/guest party and rest others are passive parties/host parties. The active party is responsible for computing training loss and generating key pairs to preserve privacy.

\noindent The vertical federated learning problem can be defined in a more formal manner. Let \begin{math} \{(x_{i},y_{i}),i=1,2,..,n\} \end{math} be a dataset where $x_{i}$ $\epsilon$ $\mathbb{R}^{d}$ and $y_{i}$ denotes the feature vector and output labels respectively. The feature dimension is represented by $d$. In case of a VFL setting, the dataset is partitioned vertically across $M$ parties/clients where each of the $M$ parties possesses a disjoint subset of features vector $x_{[i,m]}$($x_{[i,m]}$ $\epsilon$ $\mathbb{R}^{d_{m}}$$(i=1,2,..,M))$, where $d_{m}$ is the feature dimension of the $m$th party and $\sum_{m=1}^{M} d_{m} = d$. Similarly, \begin{math} \Theta = (\theta_{1}, \theta_{2},..,\theta_{M}) \end{math} can be defined where, \begin{math} \theta_{m} \epsilon \mathbb{R}^{d_{m}} \end{math} denotes the model parameter of the $m$-th party. Ideally, in VFL one of the collaborating parties is assumed to have the data labels. The party possessing label information is referred to as active party and the ones without label information as passive party. Considering the M-th party as the active party that holds the label information $y_{[i,M]}$($y_{[i,M]}$ $\epsilon$ $\mathbb{R})$, the following function is minimized while training models in a VFL setting.
\begin{equation}
\centering
    \mathcal{L}(\Theta) = \frac{1}{n}\sum_{i=1}^{n}l(\sum_{m=1}^{M} x_{[i,m]}\theta_{i}, y_{[i,M]}) + \lambda R(\Theta)
\label{eqn}
\end{equation}
Here $l(.)$ and $R(.)$ denote loss function and regularizer respectively while \begin{math}\lambda\end{math} is the tuning parameter. VFL has been known to be used in solving regression \citep{he2021secure} and classification \citep{feng2020multi} problems. The Algorithm \ref{algorithm_vfl} \citep{zhu2021federated} shows the process of solving a logistic regression problem in a VFL setting.
\begin{algorithm}[h!]
	\caption{Vertical Federated Logistic Regression} 
	\begin{algorithmic}[1]
    \State $M$ $\rightarrow$ Number of Clients
    \State $T$ $\rightarrow$ Number of Communication Rounds
    \State $B$ $\rightarrow$ Number of Batches
     \State Training Data $X = \{X^1, X^2,.., X^M\}$
    \State Client $1$ $\rightarrow$ Active Party/Guest Client
    \State Client $(2..M)$ $\rightarrow$ Passive Party/Host Client
    \State Initialize local model ${\theta_{0}}^{m}, m$ $\epsilon$  $(1..M)$
    
		\For {$each$ $communication$ $round$ $t=1,2,\ldots, T$}
			\For {$each$ $batch$ $data$ ${X_{b}}^{m}$ $\epsilon$ $({X_{1}}^{m}, {X_{2}}^{m},..,{X_{B}}^{m})$}
                \For {$each$ $Client$ $m=1,2,..M$}
				\State Compute ${z_{b}}^{m}$ = ${X_{b}^{m}}$${\theta_{t}}^{m}$
                    \If {$m \neq 1$}
                        \State send ${z_{b}}^{m}$ to Client $1$ 
                    \EndIf
                \EndFor
				\State Compute $\hat{y}_{b}$ = $\sum_{m=1}^{M} {z_{b}}^{m}$ and $L(y_{b},\hat{y}_{b})$ on Client $1$
                \State Compute $\frac{\partial L}{\partial z_{b}}$ on Client 1
                \State Send $\frac{\partial L}{\partial {z_{b}^{m}}}$ to Client $m, m \epsilon (2,k)$
                \For {$each$ $Client$ $m=1,2,..M$}
                \State ${{\theta}_{t}}^{m} \leftarrow {\theta_{t}}^{m} - \eta\frac{\partial L}{\partial {z_{b}^{m}}}\frac{\partial {z_{b}^{m}}}{\partial {\theta}_{t}^{m}}$
                \EndFor
			\EndFor
		\EndFor
	\end{algorithmic} 
 \label{algorithm_vfl}
\end{algorithm}
\noindent VFL has also been proposed for other machine learning algorithms such as linear regression \citep{gascon2017privacy, kikuchi2018privacy}, decision trees \citep{khodaparast2018privacy}, random forests \citep{liu2020federated} and neural networks \citep{zhou2020vertically} etc.\\\\
\textbf{Extensions for Vertical Federated Learning: }There has been some existing research conducted focusing on implementing the idea of VFL in extended scenarios. For instance, current VFL systems are developed on the assumption that the labels are possessed by only one client i.e. the active party. However, there might be practical cases where multiple collaborating clients possess labels which arises the need to apply VFL in a modified manner. The Multi-VFL proposed by \cite{mugunthan2021multi} makes use of split learning in a scenario where there are multiple data and label owners. Here, forward propagation is performed by the data owners on their corresponding partial models until the cut layer and then, their activations are sent to the label owners. These activations are concatenated by the label owners in order to complete their forward propagation. Subsequently, the losses are computed and back propagation is performed to compute the gradients. The gradients are then send back to the data owners who are supposed to use them for completing their back propagation. Moreover, \cite{zhu2021pivodl} proposed a secure vertical FL framework PIVODL, to train gradient boosting decision trees (GBDTs) with data labels distributed on multiple devices. PIVODL presents a setting of training XGBoost decision tree models in VFL where each of the participating client holds parts of data labels that cannot be disclosed to others during the training process. A similar approach is also observed in \citep{9973381} in order to deal with VFL when labels are distributed among multiple parties.

\subsection{Improvements to Vertical Federated Learning}
\begin{figure}[h!]
\centering
\includegraphics[width=\textwidth]{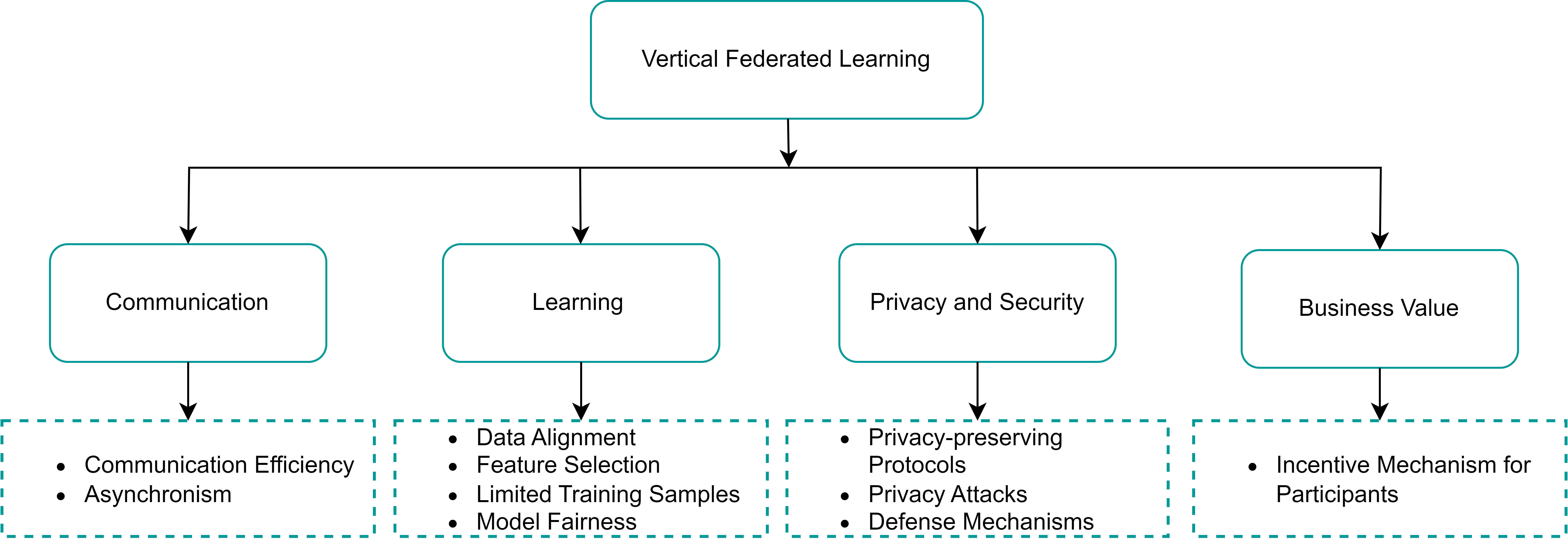}
\caption{Categorization of Vertical Federated Learning Literature}
\label{category}
\end{figure}

\begin{figure}[ht!]
\centering
\includegraphics[width=\textwidth]{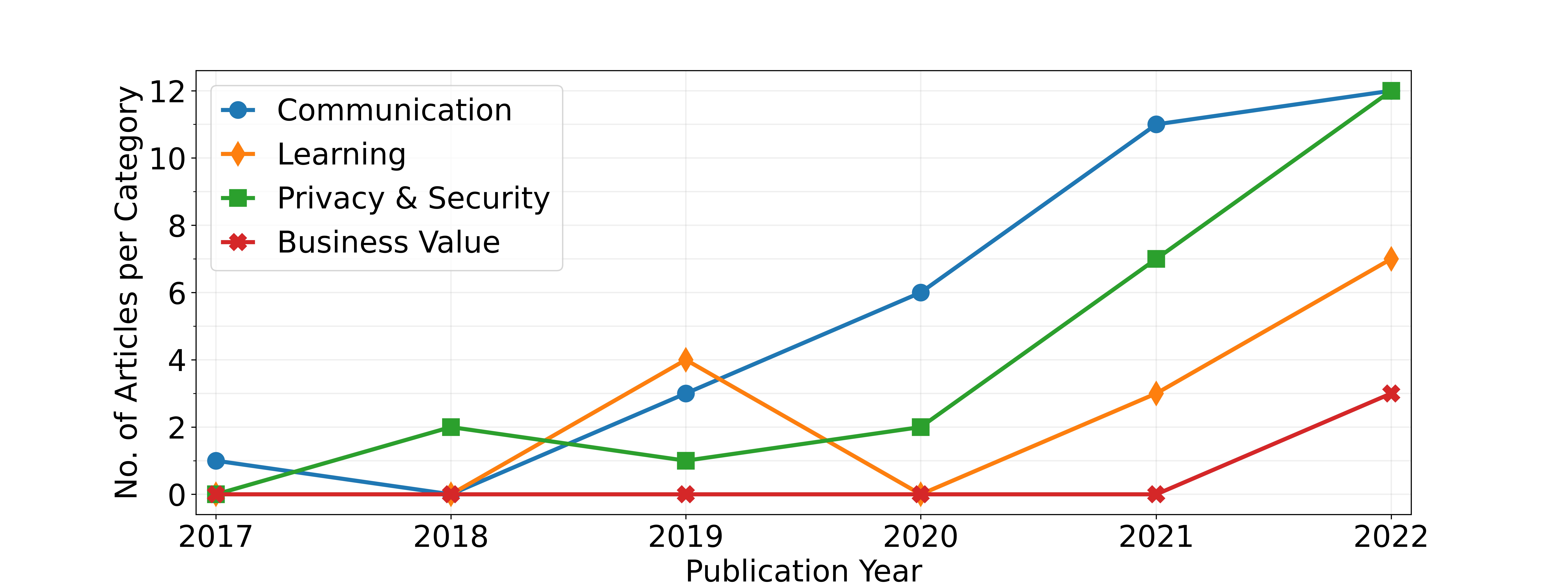}
\caption{Observed Trend in Selected Studies in VFL per Category}
\label{category_trend}
\end{figure}

The existing literature on vertical federated learning can be categorized into four groups; communication, learning, privacy \& security and business value. A brief overview of recent research on VFL in these fields is presented in Figure \ref{category}. On the other hand, Figure \ref{category_trend} depicts the trend of the selected studies per category over the years. 
\subsubsection{Communication}
Efficient communication and asynchronous updates are critical in VFL to achieve efficient and scalable distributed machine learning. As VFL involves multiple parties collaborating to train a model while keeping their data private, effective communication protocols are necessary to ensure that parties can exchange information and update model parameters accurately and efficiently. Asynchronous updates enable parties to update model parameters independently, reducing the waiting time for other parties to complete their updates. This approach can improve the efficiency of the training process and allow parties to work independently, making VFL more scalable. This section discusses some of the existing approaches found in the studies to deal with communication efficiency and asynchronism in VFL settings.
\paragraph{Communication Efficiency}
When following the conventional VFL approach, each of the passive/host clients share their updated gradients or intermediate results with the active/guest client during every training iteration. The total communication for each client can significantly increase over the course of hundreds or thousands of training iterations for very large data sets. As a result, the learning process might become inefficient due to communication cost and bandwidth constraints. Some existing researches \citep{liu2022fedbcd,yang2019quasi, xu2021fedv, xie2022improving, fu2022towards} deals with the communication overhead problem in VFL by reducing the number of local model updates during training. A Federated Stochastic Block Coordinate Descent (FedBCD) algorithm was proposed by \cite{liu2022fedbcd} for vertically partitioned data, wherein each party performs multiple local updates before each communication in order to reduce the number of client communication rounds significantly. Furthermore, Quasi-Newton method based vertical federated learning systems \citep{yang2019quasi, wenjie2021vertical} proposed where descent steps scaled by approximate Hessian information are performed leading to faster convergence than Stochastic Gradient Descent (SGD)-based methods This allowed significant reduction in the number of communication rounds. \cite{zhang2022adaptive} proposed an algorithm to minimize the training time of VFL by adaptive selection of the number of local training updates for each party.\\

\begin{table*}[t!]
\begin{adjustbox}{width=\textwidth}
\begin{tabular}{p{2.5cm}p{4cm}p{7cm}p{3.5cm}p{5cm}}
\hline
& \multicolumn{1}{c}{Article} & \multicolumn{1}{c}{Method} & \multicolumn{1}{c}{Model} & \multicolumn{1}{c}{Dataset} \\\hline
\multirow{6}{2.5cm}{Modification in Local Updates} & \citep{liu2022fedbcd} & Stochastic Block Coordinate Descent with multiple update of local models & Logistic Regression, Neural Network & MIMIC-III, NUS-WIDE, MNIST, Default-Credit \\
& \citep{yang2019quasi}, \citep{wenjie2021vertical} & Quasi-Newton Method & Logistic Regression & Default-Credit \\
& \citep{xu2021fedv} & Eliminates need for  peer to peer communication among clients by using functional encryption schemes & Linear regression, Logistic regression, linear SVM & Website phishing, Ionosphere, Landsat satellite, Optical recognition of  handwritten digits, MNIST \\
& \citep{xie2022improving} & Allowed multiple local updates in each round by using alternating direction of multipliers & Convolutional Neural Network & MNIST, CIFAR-10, NUS-WIDE, ModelNet40 \\
& \citep{fu2022towards} & Cache enabled local  updates at each client & Neural Network & $Criteo^{5}$, $zu^{6}$ \\
& \citep{zhang2022adaptive} & Adaptive selection of  local updates & Logistic Regression, Neural Network & a9a, MNIST, Citeseer \\\hline
\multirow{8}{2.5cm}{Compression} & \citep{castiglia2022compressed} & Arbitrary compression scheme on gradients of local models & Neural Network & MIMIC-III, CIFAR-10, ModelNet40 \\
& \citep{yang2021model} & Transmission of  selective gradients after compression & Logistic Regression &
  Default Credit \\
& \citep{li2020efficient} & Double-end sparse compression on local models & Logistic Regression, Neural Network & Default Credit, Insurance claim dataset \\
& \citep{khan2022communication} & Compression on local data using Autoencoders & Logistic Regression, SVM & Adult income, Wine-quality, Breast cancer, Rice MSC \\
& \citep{ratadiya2020decentralized} & Compression on local data using Autoencoders & Logistic Regression & Bank loan dataset \\
& \citep{cha2021implementing} & Compression on local data using Autoencoders & Neural Network & Adult income, Vestibular Schwannoma Dataset, The eICU Collaborative Research Database \\
& \citep{sha2021feature} & Compression on local data containing images using feature maps & Neural Network & CIFAR-10, CIFAR-100, CINIC-10 \\
& \citep{wu2022practical} & Compression on local data using unsupervised representation learning & Neural Network & NUS-WIDE, MNIST \\\hline
\end{tabular}
\end{adjustbox}
\caption{Approaches to Reduce Communication Overhead in VFL}
\label{table2}
\end{table*}

Another widely used strategy to achieve communication efficiency in VFL is the application of compression schemes to the data that is being shared among the clients. \cite{castiglia2022compressed} demonstrates efficiency of VFL improves when the data to be transmitted such as gradients of clients are compressed (using quantization or sparsification) before sharing. Based on this idea, \cite{yang2021model} proposed a method of gradient sharing and compression in VFL where only the gradients greater than a certain threshold are selected and then compressed by each of the clients before sharing in order to reduce communication bandwidth. Similarly, \cite{li2020efficient} proposed an efficient vertical federated learning framework with gradient prediction and double-end sparse compression, where the compression occurs at the local models to reduce training time as well as transmission cost. Compression can also be directly applied on the local data of the host clients in a way that the compressed data contains relevant information of the local data. Later on these compressed local data are send to and aggregated by the guest client to train the model. Satisfactory outcomes to this approach has been observed in some of the research works where the compression of local data by extracting relevant information is done by using unsupervised techniques like Autoencoders \citep{khan2022communication, ratadiya2020decentralized, cha2021implementing}, Feature Maps \citep{sha2021feature} and Representation Learning \citep{wu2022practical}. An overview of the approaches found in the studies for reducing communication overhead in VFL has been shown in Table \ref{table2}.

\paragraph{Asynchronism}
Vertical federated learning setting involves collaboration of multiple clients having different features of a single data instance to train a machine learning model. But it is not always practical to assume that all the clients would be identical in terms of in storage, hardware, network connectivity, etc. Due to such variability among the clients there may be cases when one or more clients aren't participating in model updates at the same time typically resulting in asynchronous updates. Thus asynchronism can pose challenges in the proper functioning of VFL which is addresses by some of the research works by \cite{zhang2021asysqn, gu2021privacy, wei2021privacy}. A vertical asynchronous FL scheme was proposed \citep{zhang2021secure} incorporating a backward updating mechanism and a bi-level asynchronous parallel architecture. The two level parallel architecture: the inner level between active (available to share gradients) clients  and the intra level within each client. The updates at both the levels are performed asynchronously which improves the efficiency and scalability. Moreover, \cite{chen2020vafl} solved vertical FL in an asynchronous manner by allowing each client to run stochastic gradient algorithms without coordination of other clients. Thus, temporary inactivity of any client does not pose any problem in the overall training.
 
\subsubsection{Learning}
Some learning approaches that has been used to address specific challenges in vertical federated learning have been discussed below:
\paragraph{Data Alignment}
Most of the research on VFL is conducted assuming that all the participating clients possess exactly same number of samples but different features. However, in real world scenario, this assumption may not be suitable as clients may not have identical records of data. The necessity to determine the common set of samples among the clients arises but also the aspect of privacy has to be considered such that no raw data is revealed. A common technique to solve this problem is Private Set Intersection (PSI) \citep{freedman2004efficient, huang2012private, cristofaro2010practical} which is a privacy-preserving technique that is commonly used in vertical federated learning to enable parties to perform joint analysis on their data without revealing their private data. PSI allows two parties to compute the intersection of their private datasets without revealing the contents of the datasets. The basic idea behind PSI is that each party privately computes a hash function on their dataset and shares the hashes with the other party. The parties then compare their hashes to find the intersection of their datasets. The comparison is done without revealing any information about the actual data. After the intersection is computed, the parties can continue with the collaborative learning process using only the intersection of the datasets. Over the past few years, different hashing techniques for PSI have been proposed in \citep{buddhavarapu2020private, ion2020deploying, chase2020private, lu2020multi} such as Bloom Filters and Oblivious Hashing. Bloom Filters are a probabilistic data structure that allow for efficient set membership testing, while Oblivious Hashing is a cryptographic technique that enables parties to privately compute hash functions without revealing the inputs to each other. 
\paragraph{Limited Training Samples}
It is more practical to apply any of the PSI protocols before the implementation of VFL in a real world application in order to determine the common set of data records. But a crucial fact which is to be considered is that, there might not always be enough common or overlapping samples of data available among the clients. As VFL might not produce satisfactory results due to lack of sufficient data. To address this problem a solution could be to expand the training data which has to be done also in a privacy preserving manner. A data augmentation method, FedDA proposed by \cite{zhang2022data} uses generative adversarial network (GAN) to generate more overlap data by learning the features of finite overlap data and many locally existing non-overlap data among the clients. Similarly, the semi-supervised learning approach FedCVT in \citep{kang2022fedcvt} improves the performance of the VFL model with limited aligned samples by expanding the training data through estimation of representations for missing features and predicting pseudo-labels. Some other approaches to tackle the issue with limited samples include determining inferences from non-overalpping data by using Federated Transfer Learning \citep{gao2019privacy} and Oblivious Transfer \citep{ren2022improving}.
\paragraph{Feature Selection}
To improve accuracy and training time of machine learning models, feature selection is a widely used strategy. It refers to the is the practice of choosing a subset of relevant features (predictors and variables) for use in a model construction. Conventional feature selection methods like Principal Component Analysis are simpler to apply in HFL setting compared to VFL. Since features are distributed across multiple clients, it becomes challenging to use the typical feature selection methods. Federated PCA approaches have been proposed \citep{cheung2021vertical, cheung2022vertical} where feature selection is achieved in a VFL setting at each client end by sharing of eigen vectors and eigen values of the host clients with the guest client. Furthermore, \citep{feng2022vertical} proposed a VFL-based feature selection method that leverages deep learning models as well as complementary information from features in the same samples at multiple parties without data disclosure. A Federated Stochastic Dual-Gate based Feature Selection (FedSDGFS) approach has been described in \citep{li2023fedsdg} which efficiently approximates the probability of a feature being selected, with privacy protection through Partially Homomorphic Encryption without a trusted third-party. Other approaches like Gini-impurity based feature selection \& sparse learning-based unsupervised feature selection~\citep{zhang2022secure, feng2020multi} have also been investigated in VFL settings which resulted in better performance when compared to traditional VFL.
\paragraph{Model Fairness}
Machine learning models in some cases may manifest unexpected and erratic behaviors. These behaviors when have undesirable effects on users, the model can be deemed as “unfair” based on some criteria. The existing bias in the training data is one of the key causes of a model becoming unfair. As real-world data encodes bias on sensitive features such as age, gender, and so on, VFL models may adopt bias from data and become unfair to particular user groups \citep{wu2021fairness}. Again due to features being decentralized across different parties, applying existing fair ML methods to VFL models becomes challenging. \cite{qi2022fairvfl} have addressed this issue by proposing a FairVFL framework where unified and fair representations of samples are learned based on the decentralized features in a privacy-preserving way. In order to obtain fair representations adversarial learning has been used to eliminate bias from the data. A superior performance in training fair VFL model was achieved \citep{liu2021achieving} in which the fair learning task was modeled as a non-convex constrained optimization problem. The equivalent dual form of the optimization problem was considered and subsequently, an asynchronous gradient coordinate descent ascent algorithm was proposed to solve the dual problem.
\subsubsection{Privacy and Security}
Federated learning ensures privacy of data while federation among clients since training of models occurs locally and data never leaves their local sites. In vertical federated learning process, the federated model is trained by sharing of gradients or intermediate results among clients. However, some studies conclude that, there are still possibilities of sensitive private data being leaked through the local gradients in \citep{aono2017privacy}, and participants’ data can be inferred through a generative adversarial network during the prediction stage in VFL \citep{luo2021feature}.
\paragraph{Privacy-preserving Protocols}
According to recent studies on VFL, the widely used privacy-preserving protocols include homomorphic encryption (HE) is and differential privacy (DP).\\

Homomorphic encryption (HE) \citep{yi2014homomorphic} is a cryptography technique which allows specific types of computations to be carried out on ciphertexts and generates an encrypted result which, when decrypted, matches the result of operations performed on the plaintexts. For the purpose of encrupting intermediate results (e.g. gradients), VFL typically utilizes additively homomorphic encryption like Paillier \citep{paillier1999public}. Additively homomorphic encryption allows participants to encrypt their data with a known public key and perform computation with the encrypted data by other participants with the same public key. The encrypted data needs to be sent to the private key holder so that it can be decrypted. A secure cooperative learning framework was proposed \citep{hardy2017private} for vertically partitioned data using additional HE. The framework was evaluated to be precise as the non-private solution of centralized data. Moreover, it scaled to problems with large number of samples and features. Similarly, \citep{dai2021vertical} proposed a privacy-preserving DNN model training scheme based on homomorphic encryption is for vertically segmented datasets. Moreover, Several other studies \citep{yang2019parallel, yang2019quasi, ou2020homomorphic} also have used HE as a privacy preserving protocol while proposing vertical federated learning approaches.\\

Differential privacy (DP) is a privacy-preserving protocol for bounding and quantifying the privacy leakage of sensitive data when performing learning tasks. It relies on adding noise to original data or training results to protect privacy. Too much noise can degrade the model's performance, while too less data can breach privacy. Hence, a balance between performance and privacy has to be achieved here. Wang et al. \citep{wang2020hybrid} designed a DP-based privacy-preserving algorithm to ensure the data confidentiality of VFL participants. The algorithm, when implemented, was quantitatively and qualitatively similar to generalized linear models, learned in an idealized non-private VFL setting. A multiparty learning framework for vertically partitioned datasets proposed in \citep{xu2021achieving} achieves differential privacy of the released model by incorporating noise to the objective function. In this case, the framework requires only a single round of noise addition and secure aggregation. In addition to using DP during model training, it can also be used during the model evaluation phase in VFL since there is also possibility of leaking private label information. Sun et al. proposed two evaluation algorithms in \citep{sun2022differentially} that accurately computes the widely used AUC (area under curve) metric when using label DP \citep{ghazi2021deep} in VFL.

\paragraph{Privacy Attacks in VFL}
Attacks in federated learning refer to malicious attempts by an attacker to manipulate or compromise the integrity of the federated learning process. Label inference attacks and feature inference attacks are the most commonly explored adversarial attacks in VFL. A label inference attack in vertical federated learning is a type of privacy attack where a malicious participant tries to infer the labels (i.e., the output values) associated with the training data of other participants, with the aim of obtaining sensitive information. Because no raw data is shared between the two parties, VFL initially appears to be private. At the end of a passive party, a considerable amount of information still exists in the cut layer embedding that can be used by the active party to leak raw data \citep{fu2022label}. Fu et. al in \citep{fu2022label} explored the possibilities of inferring labels exploiting the gradient update mechanism in VFL and as well as inferring labels from the bottom model trained locally by each participant to embed the input features to a latent space, which avoids the raw features being directly sent to the server. \citep{zou2022defending}  also showed that private labels could be reconstructed with high accuracy by training a gradient inversion model even with HE-communication. Moreover, \citep{sun2022label} proposed a label inference method where it was possible to steal private labels effectively from the shared intermediate embedding even though label differential privacy and gradients perturbation were applied.
\\\\
On the other hand, a feature inference attack is a type of privacy attack that aims to infer private information about a party's data based on the model's output. In vertical federated learning, an attacker may try to infer a party's private features by analyzing the model's output on the shared features. For example, an attacker may try to infer a person's medical condition based on their zip code and medical procedures, which are shared between different parties. \citep{ye2022feature} proved theoretically and experimentally that, it is possible to perform a reconstruction attack on the features of the parties by applying a robust search-based attack algorithm unless the feature dimension is considerably large enough. Furthermore, \citep{luo2021feature} proposed a general feature inference attack which learns the correlations between the features of the adversary and the attack target’s features based on multiple predictions accumulated by the adversary. 
\paragraph{Defense Mechanisms}
Defense mechanisms are crucial for vertical federated learning because this approach involves training machine learning models on data that is distributed across different entities or organizations. This means that the data is often sensitive and private, and its exposure could lead to significant risks, such as identity theft, financial fraud, or discrimination. Mainstream defense mechanisms like adding noise to gradients \citep{zhu2019deep}, gradient compression \citep{lin2017deep}, and randomly pruning part of the gradients to zero \citep{shokri2015privacy} are not able to mitigate possible label information leakage in VFL settings \citep{fu2022label}. \citep{sun2022label} proposed adding an additional optimization goal at the label party and limiting the label stealing ability of the adversary by minimizing the distance correlation between the intermediate embedding and corresponding private labels. Similarly, a dispersed training framework was also been proposed in \citep{wang2022beyond} where secret sharing had been used in order to break the correlation between the bottom model and the training data. In addition, \citep{wang2022beyond} also described a customized model aggregation method such that the shared model can be privately combined, and as well as the training accuracy in ensured by the linearity of secret sharing schemes. Furthermore, \citep{zou2022defending} demonstrated that using confusional autoencoder (CAE); a technique based on autoencoder and entropy regularization and its variant DiscreteSGD-enhanced CAE could successfully block label inference attacks without significantly hampering the accuracy of the main task. 
\\\\
Additionally, \citep{qiu2022all} proposed HashVFL which is a hashing based VFL framework to deal with feature reconstruction attacks in a VFL setting.  The one way nature of hashing enables blocking of all attempts to recover data from hash codes. Efficiency of HashVFL in mitigating reconstruction attacks have also been demonstrated through experimental results.An adversarial training based framework for VFL has been designed in \citep{sun2021defending} which simulates the game between an attacker (i.e. the active party) who actively reconstructs raw input from the cut layer embedding and a defender (i.e. the passive party) who aims to prevent the input leakage. A similar approach is observed in \citep{luo2021feature} which deals with malicious attack by active party during model evaluation stage of VFL.

\subsubsection{Business Value}
The concept of fairness in FL also includes collaboration fairness which implies the incentive mechanism in a FL setting. Since FL is based on collaboration, rewarding mechanism is crucial for the allocating rewards to current and potential participants of FL. To achieve that, FL needs a fair evaluation mechanism to give agents reasonable rewards. Moreover, to analyse business value of VFL in real world application determing a proper incentive mechanism for the participating clients is crucial since not doing do may result in lack of motivation from the clients to collaborate. The Shapley value (SV) \citep{roth1988shapley} is a provably fair contribution valuation metric originated from cooperative game theory. The contribution of participants can be measured by Shapley Values to calculate grouped feature importance \citep{wang2019measure}. This idea can also be extended to both synchronous and asynchronous vertical federated algorithms \citep{fan2022fair}. \citep{9880795} further extended the idea of contribution measurement of participants in VFL by incorporating the data pricing model based on Stackelberg with the hosts as the leader and the guest as the follower. This approach provided a managerial guidance on revenue distribution for FL platform owners based on their contributions.

\subsection{VFL Applications}
Google initiated project in 2016 in order to establish federated learning among Android mobile users \citep{bonawitz2017practical}. The goal was to improve the keyboard input prediction quality, while at the same time ensure the security and privacy of users. In the later stage, many use cases were addressed in several surveys and studies where FL could be implemented. There has been significant efforts in designing federated learning frameworks for industrial usage but most of them are still in their development stage. Among the existing FL frameworks, only few of them support VFL either completely or partially. Table \ref{tab:frameworks} provides an overview of the existing VFL frameworks. It can be observed from the overview that, most listed frameworks (FedML \citep{he2020fedml}, FedLearner, FederatedScope) do not support a variety of ML models for implementation. Besides, few of the frameworks don't have complete privacy features incorporated and don't possess complete documentations. Thus, it can be concluded that VFL frameworks are still in its developing stage which indicates a lot of scope to work on.
\\

\begin{table*}
\centering
\begin{adjustbox}{width=\textwidth}
\begin{tabular}{lp{4cm}ccccccc}
\hline
\multicolumn{2}{l}{Framework} & FATE & FedML & PaddleFL & FedLearner & FederatedScope & Crypten & FedTree \\ \hline
\multirow{5}{*}{\rotatebox[origin=c]{90}{\textbf{\begin{tabular}{l}Model\\Support\end{tabular}}}} & Regression & \Checkmark & \Checkmark & \Checkmark & - & \Checkmark & \Checkmark & - \\
& & & & & & & & \\ 
& Neural Network & \Checkmark & - & \Checkmark & \Checkmark & - & \Checkmark & - \\
& & & & & & & & \\ 
& Tree-Based Model & \Checkmark & - & - & \Checkmark & - & - & \Checkmark \\ \hline

\multicolumn{2}{l}{Third Party Collaborator Requirement} & \Checkmark & - & \Checkmark & \Checkmark & - & \Checkmark & \Checkmark \\ \hline
\multirow{3}{*}{\rotatebox[origin=c]{90}{\textbf{Privacy}}} & Complete Privacy of Model Parameters & \Checkmark & - & \Checkmark & - & - & \Checkmark & \Checkmark \\
& & & & & & & & \\ 
 & Complete Privacy of Model Gradients & \Checkmark & \Checkmark & \Checkmark & - & \Checkmark & \Checkmark & \Checkmark \\ \hline
\multicolumn{2}{l}{Availability of Complete Documentation} & \Checkmark & - & - & - & \Checkmark & \Checkmark & \Checkmark \\ \hline
\end{tabular}
\end{adjustbox}
\caption{Overview of Federated Learning Frameworks Supporting Vertical Partitioning (Adjusted from \citep{liu2022unifed})}
\label{tab:frameworks}
\end{table*}

Federated learning is a cutting-edge modeling mechanism which is capable of training a unified machine learning model using decentralized data while maintaining privacy and security. Thus, it has a promising application in financial, healthcare and many other industries where direct aggregation of data for training models is not feasible due to concerns like data security, privacy protection and intellectual property. Most applications are focused on horizontal federated learning and it is quite difficult to observe VFL being used in real-world applications yet. However, in our literature review, we found a limited number of studies proposing and as well as implementing VFL in applications. \citep{sun2021fedio} considered a scenario involving two hospitals; Inner and Outer hospital possessing records of daily performance and clinical test of patients. Due to patients privacy regulations, it was not allowed to share raw data among the hospitals even though they had records of the same patients. Hence, the Inner- and Outer-hospital information had been bridged via vertical Federated Learning for perioperative complications prognostic prediction. In \citep{zhou20222d} a VFL scheme is developed for the purpose of human activity recognition (HAR) across a variety of different devices from multiple individual users by integrating shareable features from heterogeneous data across different devices into a full feature space. VFL has also been useful in the field of e-commerce as observed in \citep{zhang2021vertical} where a method based on clustering and latent factor model under the vertical federated recommendation system was implemented. Taking into account the diversity of a large number of different users in each participant and the complexity of the matrix factorization of the user-item matrix, the users were clustered to reduce the dimension of the matrix and improve the accuracy of user recommendations. Similarly, efficient online advertising was achieved through the application of VFL \citep{li2022vertical}. Vertical federated learning when applied to financial institutions boosted their profits by collaboration of data among them maintaining privacy \citep{liang2021methodology, efe2021vertical}. A special use case of VFL was observed in the aviation domain \citep{guo2021research} where a flight delay prediction model based on federated learning was designed by integrating horizontal and vertical federated frameworks. 

\section{Open Challenges and Future Directions}
The structured literature review has provided insights to existing approaches adapted to improve different aspects of VFL while also pointing out potential research directions by identifying the gaps in the literature, highlighting unresolved issues and noting new developments in VFL. The research directions concluded from the review are discussed as follows:
\subsection{Communication Efficiency}
The total communication and computation cost in VFL is proportional to the size of the training data. With the growing amount of data accross multiple platforms, VFL becomes more challenging due to significant increase in local model computations, updates and as well as communication cost. To tackle this problem several sufficient studies discussed earlier have been computation and communication efficient methods that reduce the communication and computation complexity. But then again data in massively increasing day by day, on the other hand computing resources of participants are not increasing at the same rate. Hence, making VFL more efficient by ensuring low communication rounds and computation cost will still remain a challenge.
\subsection{Privacy and Security}
Privacy and Security has always been a concern in FL since the assurance of the no exposure of raw data is a major motivation for data owners to participate in the federation. Researchers in their studies so far have rigorously tried to identify potential privacy leakage and malicious attacks in FL setup. However, the aspect of security in VFL is not enough explored as most privacy preserving protocols like homomorphic encryption, secret sharing and differential privacy are used in HFL scenarios. Moreover, possibilities and defense mechanisms for poison attacks; introduction of malicious data during training phase by adversary and backdoor attacks; insertion of malicious functionality into a targeted model through poisoned updates from malicious clients in VFL settings have been very limitedly explored. Hence, there is still scope for improvements in dealing with privacy and integrity leakage in VFL.
\subsection{Limited Training Data}
As a preprocessing step for VFL, the overlapping samples among the clients are determined. Often the overlapping samples are insufficient for achieving a good performance on VFL models. Expanding the training data is a solution but then again that also comes with privacy constraints since local raw data cannot be disclosed. Again not utilizing the non-overlapping samples among the participants would mean making the data useless. Since data is expensive and difficult to obtain, new approaches could be designed such that relevant information is inferred from the non-overlapping data as well. 
\subsection{Feature Selection}
Feature selection, which has been a topic of research in both methodology and practice for decades, is used in a variety of different fields like text mining, image recognition, fault diagnosis, intrusion detection, and so on. VFL has promising potential in many feature selection applications with privacy preservation. Feature selection combined with VFL would allow business from different organizations to collaboratively perform feature selection without exposing private data. Privacy-preserving feature selection in VFL has not been fully explored yet, although only a few solutions were presented (Section 3.2.2). Hence, designing efficient and effective privacy-preserving feature selection protocols for VFL could be an interesting direction for future research.
\subsection{Model Fairness}
There has been rising interest in developing fair methods for machine learning \citep{yucer2020exploring}. In practical VFL setups, the aspects like data being distributed across multiple platforms and asynchronous parallelized updates makes model fairness even more challenging to enhance. However, such concerns have been less addressed in vertical federated learning settings. The application of fair VFL methods for ensuring bias free model training is an open opportunity for future vertical federated learning research. It is particularly important as VFL has potential to be implemented in applications involving real populations of users without knowledge of their sensitive identities.
\subsection{Incentive Mechanism}
Motivating data owners to participate in a data federation is a significant challenge in federated learning. Even in vertically federated setups, it is essential to encourage more qualified clients to participate. In order to achieve this goal, incentive schemes are to be designed which are able to fairly share the profits generated due to the collaboration among participants. This cannot be done without first establishing a mechanism for assessing each data owner's contribution to the federated model. Despite the fact that several works have focused on the design of incentive mechanisms for vertical federated learning, one crucial aspect, i.e. security, has been overlooked. Even while evaluating contributions of participants, it is important that privacy is preserved from all ends by ensuring no exposure of data. Improvised incentive mechanism could be designed which not only deals with distribution of profit but also penalizes the participants in case  erroneous data is provided by them misleading the federated learning process. Furthermore, almost all studies dealing with the contribution measurement of parties in VFL, have utilized the cooperative game theory concept of Shapley values to calculate participant contribution. However, calculating Shapley values is computationally expensive due to consideration of all possible coalitions. In this case, other cooperative game theory concepts can also be experimented with in context of VFL settings. To design a fair and effective incentive mechanism, in addition to contribution measurement, proper selection of participants in VFL settings \citep{jiang2022vf} is also crucial such that the best parties are only collaborated with to improve the global model. Therefore, participant selection in VFL is an intriguing research direction.
\subsection{Explainability}
An area of study that has gained a great deal of attention recently is explainable artificial intelligence (XAI). Models must be explainable in high stake applications like healthcare, when there is a strong need to justify decisions made. The same applies in case of VFL as well. VFL models must be explainable, especially when dealing with sensitive data. In VFL models, each party's data is kept private and is not accessible to third parties for analysis. For the deployment of VFL, it is crucial to interpret models while making sure the data is stored locally only. Explainability of VFL models may also be attributed to potential privacy violations from unintended data leaks. Thus, there is a potential scope for future research for explainable AI in context of VFL where a trade-off between explainability and privacy is achieved.

\section{Conclusion}\label{sec13}

With the advancement of big data and artificial intelligence, the expectations of privacy are becoming increasingly stringent. As a result, federated learning, a novel solution to cross-platform privacy protection, was developed. Apart from privacy, FL now needs to deal with a number of other challenges when applied to data partitioned in various formats. While most studies focus on FL on horizontally partitioned data, this review aims to provide researchers in FL domain with the current state-of-the-art in vertical federated learning. We not only mentioned the challenges observed in VFL in our study, but also clustered them into four categories; communication, learning, privacy \& security and business value. The study also discussed the approaches adopted by earlier studies to overcome these challenges and looked into the applicability of VFL in real-world scenarios. Finally, a set of eight prospective future directions for research in this domain have been identified.  

\backmatter

\bmhead{Supplementary information}
Not applicable.




\section*{Declarations}
\begin{itemize}
\item Funding: Not Applicable
\item Conflict of interest: The authors declare no conflict of interest. 
\end{itemize}

\bibliography{sn-bibliography}


\end{document}